\documentclass{article}
\usepackage{acra}
\usepackage{subfigure}
\usepackage{siunitx}
\usepackage{booktabs}
\usepackage{graphicx}
\usepackage{csquotes}
\usepackage{soul}
\usepackage{algpseudocode}
\usepackage{amsmath,amssymb}
\usepackage{algorithm2e}
\usepackage{float}
\usepackage{ragged2e}
\usepackage{adjustbox}
\usepackage{cite}

\usepackage{url}
\usepackage{multirow}
\usepackage{array}

\newtheorem{Def}{Definition}[section]

\newcounter{example}[section]
\newenvironment{example}[1][]{\refstepcounter{example}\par\medskip
   \noindent \textbf{Example~\theexample. #1} \rmfamily}{\medskip}
   
\DeclareMathOperator{\rk}{rk}

\begin{document}

\title{\vspace{-1.2cm} Learning Behavioral Representations of Human Mobility \thanks{The final version of this work will appear in the proceedings of ACM SIGSPATIAL 2020 with DOI: 10.1145/3397536.3422255. Please refer to the final version for future citations.}}

\author{
\begin{tabular}{*{2}{>{\centering}p{.5\textwidth}}}
 \multicolumn{2}{c}{ }\tabularnewline
\large Maria Luisa Damiani & \large Andrea Acquaviva  \tabularnewline
\small\url{maria.damiani@unimi.it} & \small\url{andrea.acquaviva1@studenti.unimi.it} \tabularnewline 
 \multicolumn{2}{c}{ }\tabularnewline
 \large Fatima Hachem & \large Matteo Rossini \tabularnewline
\small\url{fatme.hachem@unimi.it} &\small\url{matteo.rossini@unimi.it} \tabularnewline
\multicolumn{2}{c}{ }\tabularnewline
\multicolumn{2}{c}{Department of Computer Science}\tabularnewline
\multicolumn{2}{c}{University of Milan, Italy}\tabularnewline
\end{tabular}
}



\maketitle

\vspace*{0.3cm}
\begin{abstract}
		 In this paper, we investigate the suitability of state-of-the-art representation learning methods to the analysis of  behavioral similarity of moving individuals, based on CDR trajectories. The core of the contribution is a novel methodological framework, \emph{mob2vec}, centered on the combined use of a  recent symbolic trajectory segmentation method for the removal of noise, a novel trajectory generalization method incorporating behavioral information, and an unsupervised technique for the  learning of  vector representations  from sequential data. \emph{mob2vec} is the result of an empirical study   conducted on real  CDR data through an extensive experimentation. As a result, it is shown that \emph{mob2vec} generates vector representations of CDR trajectories  in low dimensional spaces which preserve  the similarity of the  mobility behavior of individuals.
\end{abstract}

\section{Introduction}
Core resources for the study of human mobility are the datasets of Call Detail Records (CDR) \cite{2013physica}. CDRs report the communication activities of mobile communication subscribers as series
of geo-referenced events, i.e., voice call start/end, text message, data upload/download, collected by
mobile operators for billing purposes \cite{DamianiHQG19}. We refer to a series of temporally ordered CDRs associated to an individual as CDR trajectory.   
Abstractly, CDR trajectories can be represented as sequences  of timestamped symbolic locations (i.e. geo-referenced symbols),  for example,  base station identifiers. Therefore, given a set of symbolic locations $L$, a  CDR trajectory $T$ can be represented by the  sequence:
$$T=(l_1,t_1),\ldots,(l_n,t_n)\quad\hbox{with}\quad l_i \in L $$

CDR trajectories are complex data, encompassing both the spatio-temporal and the textual dimensions,  irregularly sampled, spatially sparse, and noisy. Yet, CDR data offer unique opportunities for the study of human mobility because of the large base of  users  monitored in their daily life.

Seminal research on human mobility modeling has shown  that people  exhibit a high degree of regularity in their motion \cite{barabasi2008,barabasi2010}. In particular, individuals visit recurrently few locations and only sometimes divert from their habitual trips to visit new locations \cite{barabasi2008}. 
While regularity is a well established characteristic of human mobility, on which there is a broad consensus in literature, a challenging question is how to compare the mobility behavior of  individuals and determine whether and to what extent those behaviors are similar. As an example, consider 
users  $u_1$ and $u_2$, and suppose that $u_1$ travels daily from location $l_1$ to $l_2$ and back to $l_1$, thus exhibiting the pattern  $l_1\rightarrow l_2 \rightarrow l_1 $, while user $u_2$ follows the pattern $l_3 \rightarrow l_4 \rightarrow l_3$; moreover assume that both users  very rarely  frequent  other  locations. The intuition is that the mobility  behavior  is  similar,  regardless of the actual locations where they stay, e.g., users visit a similar  number of locations in a similar way. We refer to this form of similarity as \emph{behavioral similarity}. 

 Behavioral similarity is hard to define. Besides, it  does not depend on the  specificity of locations and thus cannot be approached using  classic similarity metrics, e.g.  \cite{lcss2018}, sequential pattern mining techniques, e.g. \cite{fournier2017frequentmining, mooney2013frequentmining},
recent techniques for the representation learning \cite{Bengio2013} of user activities graphs, e.g. \cite{kdd2019,kdd2018}.
To tackle the problem, a viable approach  is to characterize the individual behavior by measuring selected mobility features. 
For example, a   metric  widely used  by the human mobility research community, is the radius of gyration $R_g$ (and its variants, e.g. \cite{barbosa2018,pappalardo2015gyration}) measuring the distance from the trajectory center of mass. 
High values of $R_g$ are  interpreted as high user's propensity to mobility.  Another kind of measure is \emph{true location diversity}, an entropy-based  metric that  we have recently proposed to  quantify the heterogeneity of  locations in symbolic  trajectories \cite{DamianiHQG19}.  
In general, however, the approaches grounded on the choice of metrics present an important shortcoming, namely it is not clear which  mobility features are really significant to characterize the mobility behavior. Consequently any metric results arbitrary or partial. 
%

To deal with the question, we explore a novel approach  built on representation learning, applied to generalized trajectories incorporating behavioral information. 
For this study, we consider CDR trajectories limited to their symbolic and temporal dimensions. 
The idea is to investigate the suitability of techniques  drawn from NLP, i.e.  word2vec \cite{word2vec2} and Paragraph Vector \cite{ParagraphVector1,ParagraphVector2}, to generate vector representation of CDR trajectories. 
The problem can be broadly formulated as follows: given an n-dimensional vector space $\mathbb{R}^n$ and a dataset $D$ of  trajectories, we search for a mapping $f\colon D\rightarrow\mathbb{R}^n$ capable of preserving mobility behavior similarity. $\mathbb{R}^n$ is the \emph{embedding space}, the vector $v=[x_1,\ldots,x_n]$ the trajectory representation or \emph{trajectory  embedding}. 
CDR trajectories have peculiar features which raise important challenges: 
the dictionary of symbols is small (a few hundred symbols), trajectories are of different length, symbols are temporally annotated, data are noisy and contain a large number of rarely frequented locations. Therefore, the representation learning  techniques used in NLP cannot be straightforwardly applied to CDR trajectories. On the other hand, existing techniques for the vector  representation  of trajectories,  such as \cite{t2vec,t2vec-explained},  focus on  spatial trajectories, thus overlooking the discrete, symbolic dimension of movement, while, by contrast, techniques focusing on symbolic locations typically  organize the set of locations visited by users in graphs, and not in temporal sequences, e.g \cite{kdd2019,sigspatial2019}, or rather interpret behavioral similarity in terms of location similarity, e.g. \cite{sigspatial2017}.  
%

In this paper, we present an empirical approach to the generation of trajectory embeddings resulting in a methodological framework  called \emph{mob2vec}. \emph{mob2vec} is the outcome of an extensive experimentation conducted  for evaluating  different options and alternative directions.
\emph{mob2vec} is grounded on a number of key design choices, in particular:
\begin{itemize}
    \item 

\noindent
It builds on a state-of-the-art technique for the  generation of embeddings for symbolic sequences called Sqn2Vec \cite{Sqn2Vec}.
Sqn2Vec is grounded on a feedforward  Neural Network architecture and can be applied to datasets of sequences of arbitrary length and containing a limited number of symbols. This technique, however, does not provide any support to deal with time and noisy data.

\item 
It leverages a technique we have recently proposed for the removal of noise and extraction of significant locations from CDR and symbolic trajectories, called SeqScan-d. The resulting trajectories are called \emph{summary trajectories} \cite{DamianiHQG19}.

\item 
 Summary
 trajectories are mapped onto generalized 
 trajectories in which location names are replaced by a measure of location relevance, expressed in terms of \emph{frequency rank}.  For example, the most visited location has rank 1, the second most visited rank 2 and so on. The rank variable thus takes  ordinal values. The resulting trajectories are called \emph{rank trajectories}. 
\end{itemize}

These design choices drive the representation learning process, which  
consists of the following steps: \emph{data preparation} turns CDR trajectories into noise-free rank trajectories; \emph{model training} and \emph{trajectory embedding generation} generate  vector representations of   trajectories in a multidimensional space; finally \emph{dimensionality reduction} maps the vector representations  onto a low-dimensional space. 
%

In summary, the main contributions of our research are: 
(i) a composite  methodology integrating representation learning, advanced trajectory segmentation techniques, and trajectory generalization strategies;
(ii) a detailed workflow for  the generation of trajectory embeddings;
(iii) an evaluation metric for quantifying the goodness of trajectory embeddings;
(iv)
a validation strategy to assess the similarity of trajectories based on their vector representation. 

The \emph{mob2vec} methodology has been tested and evaluated on a dataset of 17000+ CDR trajectories provided by the NPTLab of the University of Milan. The dataset  is comparable in size with the datasets  used in the Sqn2Vec project.

The rest of the paper is organized as follows: Section 2 provides background knowledge on  Sqn2Vec and SeqScan-d techniques; Section 3 introduces the  CDR dataset, while Section 4 the \emph{mob2vec} framework; Section 5 and 6 report two different classes of experiments.  Section 7 reports some conclusive considerations.

\section{Background knowledge}
In this section, we overview  key features of the two major  components used in \emph{mob2vec}, i.e. Sqn2Vec and SeqScan-d.

\subsection{Sqn2Vec}
Sqn2Vec \cite{Sqn2Vec} is an unsupervised method recently proposed for the learning of symbolic sequence representations and tailored to datasets with relatively small vocabularies.
Sqn2Vec builds on Paragraph Vector\footnote{Also known as doc2vec after the name of a popular implementation \cite{doc2vecImplementation}.} (PV) \cite{ParagraphVector1,ParagraphVector2}, and indirectly on word2vec \cite{word2vec2}, thus is  rooted in  the representation of variable-length pieces of text and words.

In these algorithms, the real objective of learning a representation of input data is accomplished by trying to solve an auxiliary correlated task.
The progress on this task is made by an iterative heuristic procedure that, at the end of each iteration, adjusts the relevant representations using linearly decreasing weights.
PV is presented in two variants that differ in the proposed task.
\begin{itemize}
\item \emph{Distributed Memory (PV-DM).}
By using the representations of both paragraphs and words, the task is: given a context (i.e.\ a bunch of consecutive words in a paragraph), predict the next word.
\item \emph{Distributed Bag-Of-Words (PV-DBOW).}
By only using the representation of the current paragraph, the task is to predict some of the words of which it is composed.
In more detail: at each iteration, a classification task is formed by taking the words in the chosen rolling context as positive samples, and random words drawn from the global word frequency distribution as negative samples.
This variant is conceptually simple, requires to store less data and has been demonstrated to be much faster.
\end{itemize}

PV can be applied not only to plain text, but also on all kinds of symbolic sequences.
However, the representation learned by PV is often poor when the training dataset has a small vocabulary.
This is typical of CDR datasets, in which users' trajectories are composed of just a few dozen unique symbols; a more extreme case is the application to DNA sequences, in which the vocabulary size is 4.
One of the possible explanations of the poor performance in these cases is that the context surrounding each symbol carries little information; moreover, in these algorithms the context is treated as a set rather than as a sequence, thus losing some information regarding the order in which the symbols appear.

To overcome the above difficulties, Sqn2Vec introduces the following concept: given a threshold $\delta\in[0,1]$, a \emph{sequential pattern satisfying a $\Delta$-gap constraint} (SP for short) is a sequence of symbols that occur in at least a fraction $\delta$ of the dataset sequences; these symbols can have a distance of at most $\Delta$ each with respect to the following one (see \cite{Sqn2Vec} for a formal definition).
The idea behind Sqn2Vec is to exploit the property that the size of the SP vocabulary can be -- depending on $\delta$ and $\Delta$ -- several orders of magnitude greater than that of the symbols, so that, by associating to every sequence its set of SPs, it is possible to construct another input dataset for a representation learning algorithm.
However, the original sequences cannot simply be replaced by their corresponding set of SPs, because a sequence may not contain any SP; so SPs have to be used \emph{in addition} to the original symbols, and not as a replacement of them.

Sqn2Vec is composed by two phases.
The first one is the sequential pattern discovery: fixed a threshold $\delta$ and a gap $\Delta$, a pattern mining procedure computes the set of SPs associated to each original sequence.
The second phase of Sqn2Vec is the sequence embedding learning.
Both original sequences and their sets of SPs are exploited in this phase, in one of the two following alternative ways.
\begin{itemize}
\item \emph{Sqn2Vec-SIM.}
For each sequence, its original symbols and its SPs are concatenated and used as input of a single PV-DBOW training process, in which positive and negative samples are drawn from a distribution made of both symbols and SPs.
\item \emph{Sqn2Vec-SEP.}
The embedding vectors of original sequences and SPs are learned separately in two independent PV-DBOW trainings.
The two representation spaces have the same dimensions, so that the resulting vectors can be pairwise averaged at the end of the process, thus producing a unique representation vector for each original sequence.
It can be expected that in this variant the training processes will produce more accurate results as they use homogeneous information.
\end{itemize}

\subsection{SeqScan-d}
SeqScan is a family of density-based trajectory segmentation techniques developed to extract sequences of \emph{relevant} locations (or stay regions) from trajectories, based on a relevance model, and designed to be robust against noise  
\cite{DamianiHIRMC18}.

\begin{figure}[t]
\centering
\subfigure[]{%
\includegraphics[width=3.5cm]{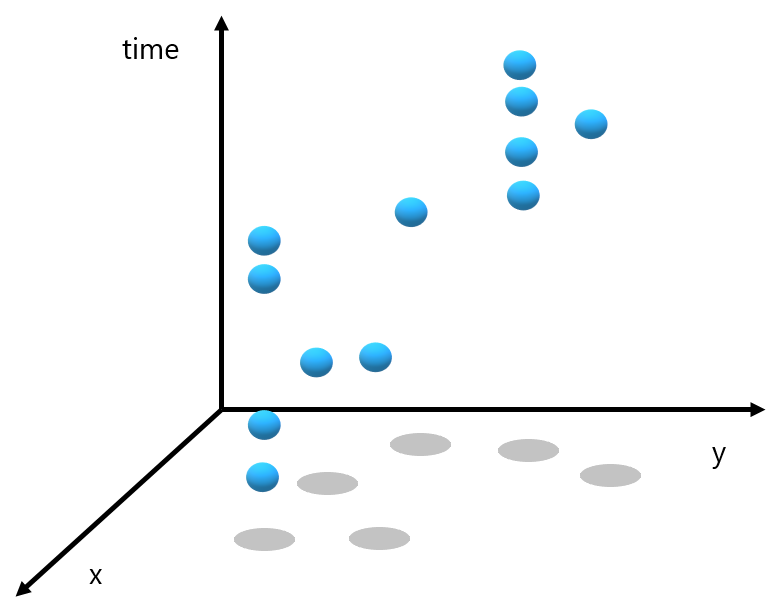}}%
\qquad
\subfigure[]{%
\includegraphics[width=3.5cm]{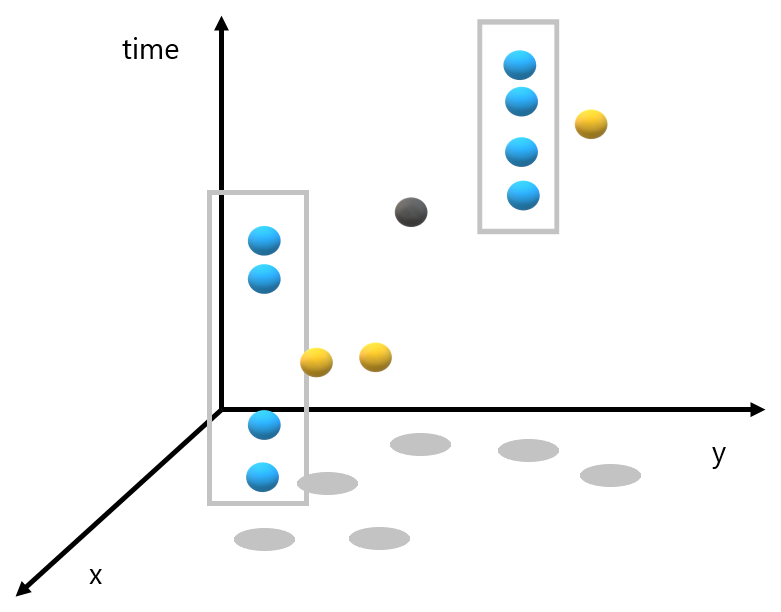}}%
\caption{(a) A CDR trajectory with
12 occurrences (in space-time).
(b) The rectangles contains two groups of occurrences for the same symbol, i.e. clusters along the temporal line. There are also 3  local noise points (colored in yellow) and one transition point (colored in black) \cite{DamianiHQG19}.}
\label{fig:trajectory3d}
\end{figure}

The outcome of the algorithm  is a series  of 
temporally ordered clusters, the \emph{summary trajectory}. Besides, two classes of noise points are identified: (a) the points temporally in between two consecutive clusters, i.e. \textit{transition}. (b) The points representing temporary absences from the cluster, referred to as \textit{local noise} \cite{Damiani2017}.
SeqScan exists in two versions, one for the segmentation of spatial trajectories; the other for the segmentation of discrete, symbolic  trajectories taking the form: $T=(l_1,t_1),\ldots,(l_n,t_n)$ where  $l_i$ is a name, or \emph{symbol}, of a dictionary \cite{DamianiHQG19}.  We refer to the discrete version used in this paper as SeqScan-d. 
%
In a symbolic space, a (SeqScan-d) cluster is a segment of maximal length satisfying the model constraints. Such segment can be concisely represented by a unique representative symbol, i.e. the symbol that, based on the model, is \emph{dominant} in the segment period. The algorithm requires two parameters $N$ and $\delta$, the former specifying the minimum number of occurrences for a symbol to be dominant in a segment; the latter the minimum cumulative amount of time that symbol appears in the segment.  Summary trajectories  can be straightforwardly represented using  the symbolic data model in \cite{tsas2015}.
SeqScan-d has been applied to the summarization of CDR trajectories. 
Figure \ref{fig:trajectory3d} exemplifies a CDR trajectory  in the discrete spatio-temporal coordinates space. 

\section{The input dataset}
We begin describing the CDR dataset used for this work. 
The dataset  is provided by a major mobile operator in Italy. The dataset covers the city of Milan plus a few surrounding districts, over a period of 67  days, from March to May 2012.  It contains 17000+ trajectories of anonymized users. The location information is given at the spatial granularity of \emph{Location Area}, where a  Location Area is a set of one or more base stations,  grouped together by the mobile operator, and univocally identified by a label \cite{DamianiHQG19}.  

\begin{figure}[t] 
	\centering
    \includegraphics[width=8.5cm]{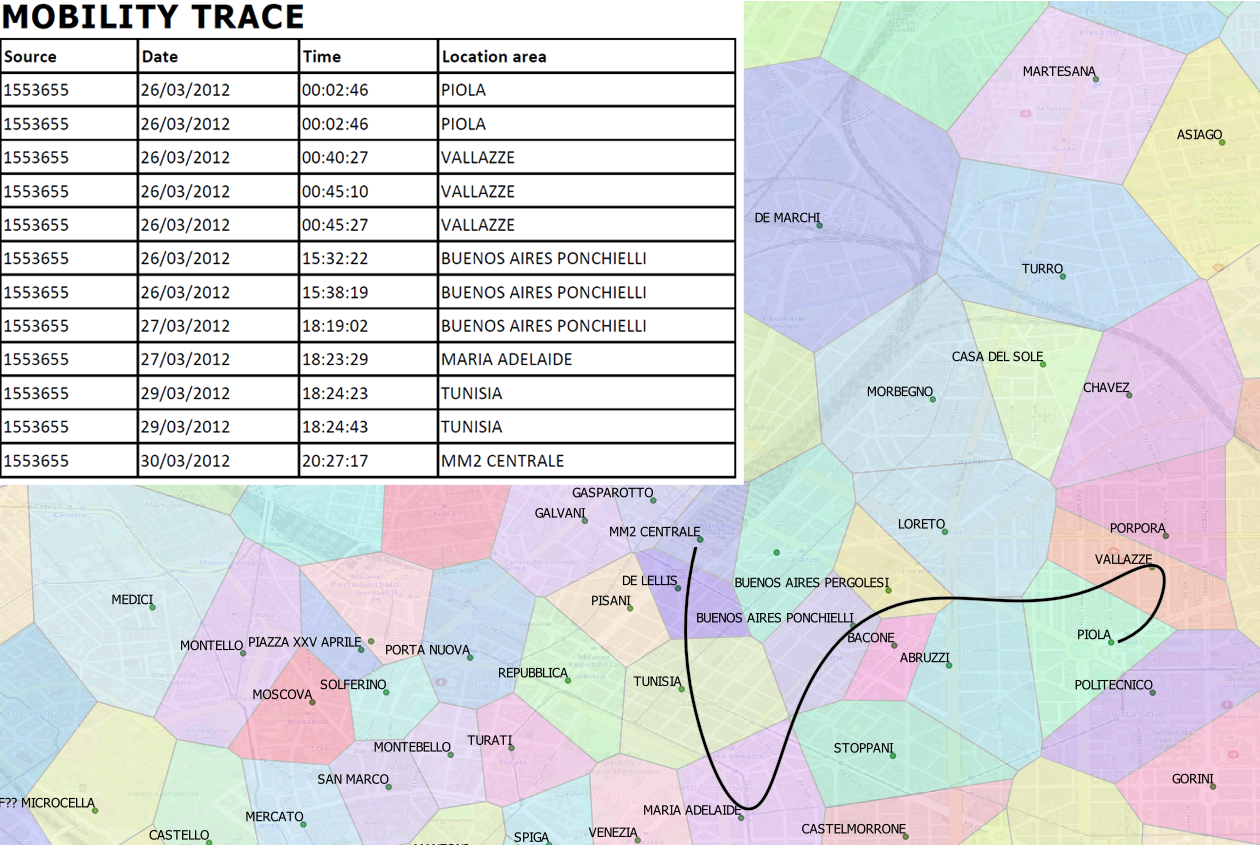}
	\caption{A fragment of CDR trajectory along with the Voronoi diagram approximating the Location Areas.}
	\label{excdr}
\end{figure}

Cells and Location Areas  coordinates are not available. However, in previous work, it was estimated that  75\% of the Location Areas in Milan are smaller
than 1 square kilometer
and concentrated downtown, whilst the largest regions, over
4 square kilometers, are in the suburbs.  Figure \ref{excdr} shows a fragment of CDR trajectory  with in background the Voronoi polygons approximating Location Areas and an example user path. A CDR  specifies  the user identifier, the timestamp and the Location Area (we omit the communication event, e.g. phone call).  The names of the Location Areas form the symbolic  space.  We refer the reader to  \cite{2018sab, 2016sab} for further details on the dataset.

\section{The \emph{mob2vec} framework}

\subsection{Overview}
The \emph{mob2vec} framework supports two  tasks: (i) generation of the representation model from the input dataset of CDR trajectories; and (ii)  representation \emph{inference} (or prediction). The former task is to populate the embedding space with the  vector representation of the  CDR trajectories of the input dataset; the inference task is to generate the representation of new trajectories, based on the learned model.

The generation of the representation model is the key task comprising four main steps  outlined as follows:


\begin{itemize}
\item \textbf{Data preparation}. Irrelevant and noisy locations are removed from input trajectories using the SeqScan-d technique. The result is a set of summary trajectories.
Summary trajectories are generalized into rank trajectories, hence  split in smaller  units, based on temporal criteria. 
The result is the \emph{source} dataset.
\item \textbf{Model training.}
Source data are pre-processed to generate the training set for the Sqn2Vec technique. The outcome of this phase is the vector representation of trajectory units.
\item \textbf{Generation of trajectory embeddings.}
The  vectors obtained at the previous step  are aggregated to form the trajectory embeddings.
\item \textbf{Dimensionality reduction.}
The multimensional embedding space is reduced to a low demensional space (e.g. 2D space), using the UMAP dimensionality reduction technique \cite{umap2018}. The embedded points in the 2D space represent thus the input CDR trajectories.
\end{itemize}

The above steps are detailed in the following. The basic notation used throughout the paper is summarized in Table  \ref{table:notation}. 
\begin{table}[t]
    \caption{Notation} 
    \label{table:notation} 
	\centering 
	\begin{tabular}{cl} 
		\toprule
		Symbol & Definition \\
		\midrule
		$T$& CDR trajectory\\
		$T_s$& Summary trajectory\\
		$T_r$& Rank trajectory\\
		$T_w^j$& Weekly (rank) trajectory for week $j$\\
	    $t_i$ & Timestamp in $T$\\
	    $I_i$ & Time interval in $T_s, T_r$\\
	    $L$ & Set of location names / symbolic locations\\
	    $R$ & Set of frequency rank values\\
	    $\rk(l_i)$, $r_i$ & Rank of location $i$\\
	    $r$ & Pearson Correlation Coefficient\\
	    $e_i$ & embedding\\
		\bottomrule 
	\end{tabular}
\end{table}

\subsection{Data preparation}
The input CDR trajectories  are processed and a number of transformations are applied to sequential data. The three key operations are:
\subsubsection{Trajectory summarization}
\begin{figure}[t]
\centering
\subfigure[]{
\label{fig:first}
\hspace{-4mm}\includegraphics[width=6cm]{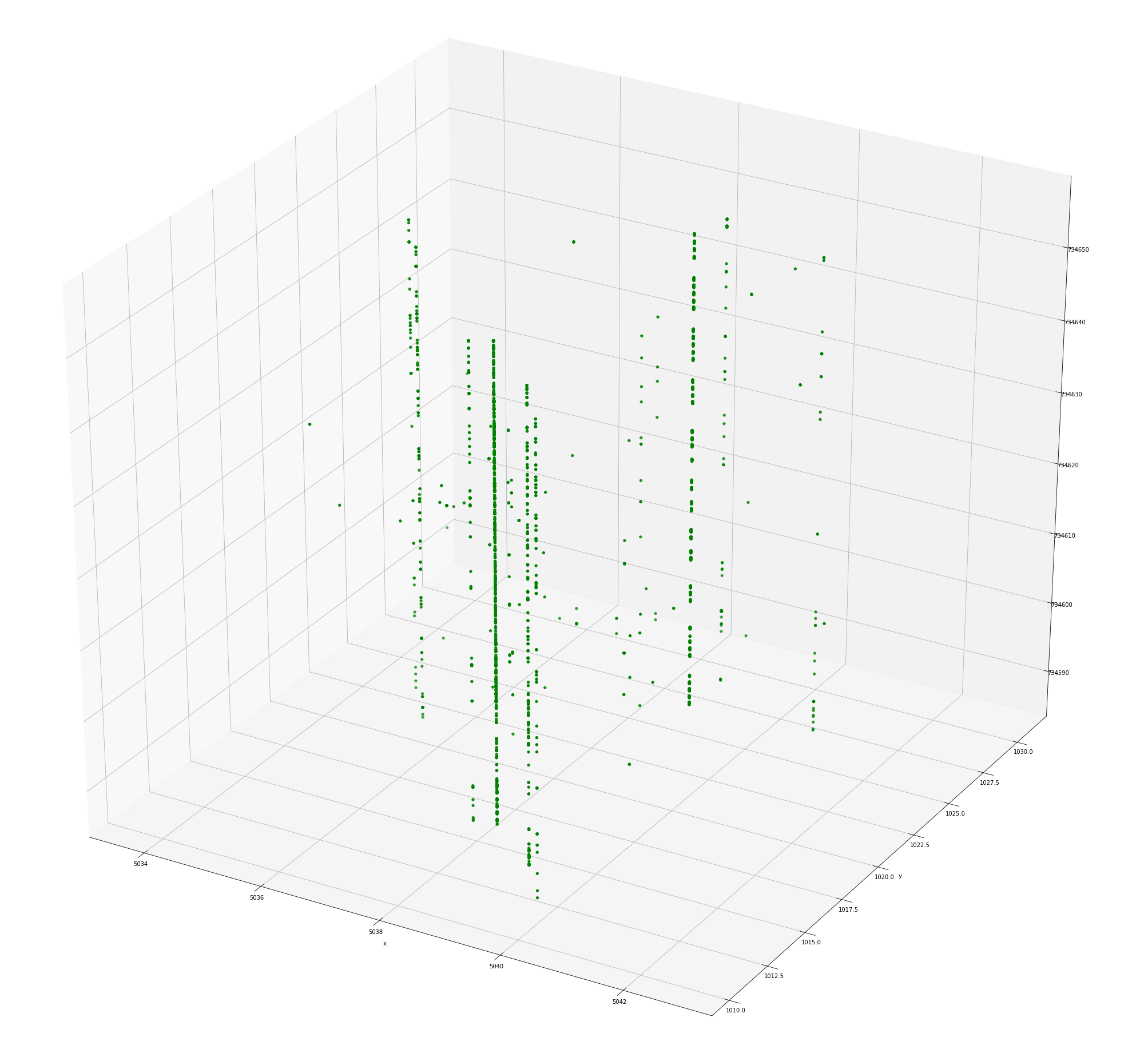}}
\subfigure[]{
\label{fig:second}
\hspace{-3mm}\includegraphics[height=4cm]{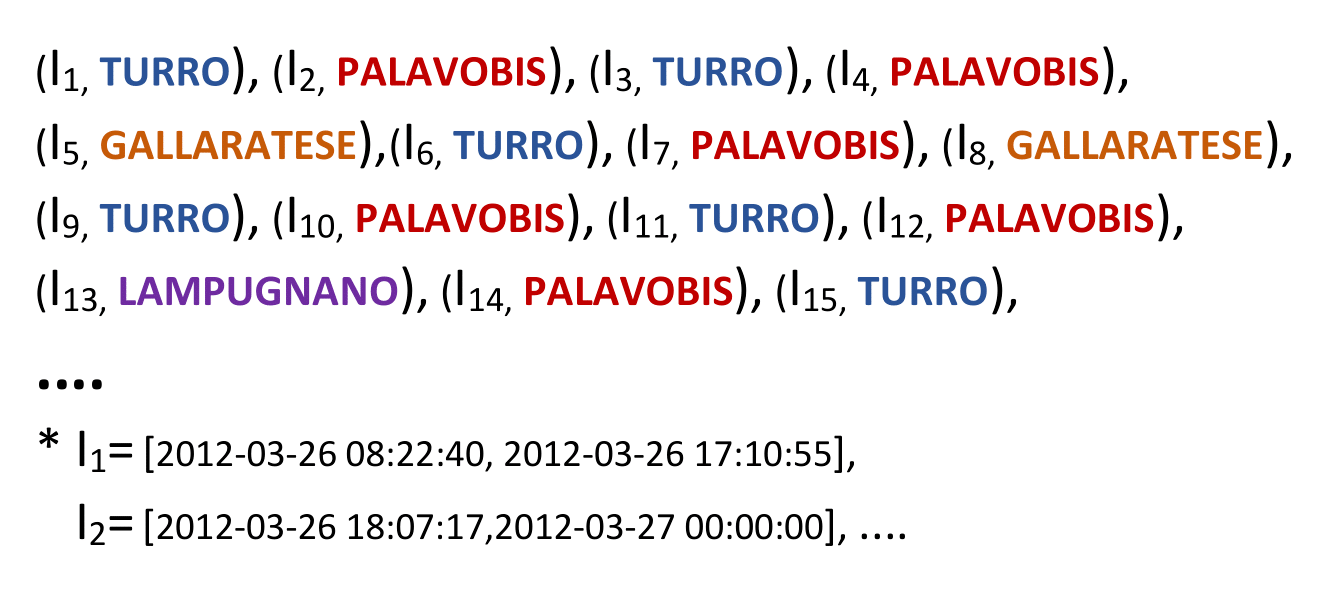}}

\caption{CDR trajectory \#1472043. (a) Graphical representation in a spatio-temporal coordinate system. The points in the plane are the Location Area centroids. (b) Portion of the corresponding summary trajectory.}
\label{es}
\end{figure}

This operation extracts from every  CDR trajectory a sequence of temporally annotated symbols.  Each symbol is representative of the series of locations appearing in a  trajectory segment compliant with the relevance model of the SeqScan-d technique. Symbols are  annotated with the temporal extent of the corresponding segment. The result is a \emph{summary trajectory}: 
$$T_s=(I_1, l_1),\ldots,(I_n, l_n)$$
where  the symbolic location $l_i \in L$ is  representative  of the i-th segment detected by the SeqScan-d algorithm, over time interval $I_i$. In the rest of the paper, we refer to the locations appearing in summary trajectories as \emph{relevant locations}.  Figure \ref{es} shows an example of summary trajectory drawn from the CDR dataset. 

The purpose of trajectory summarization is to remove noise and irrelevant locations from native trajectories, which can adversely affect the learning process.  Whe\-ther a summary trajectory is really representative of the actual movement of an individual is a question that has been tackled in recent work. In \cite{tsas2020} we show that summary trajectories preserve major statistical properties of the native CDR trajectories, specifically the distribution of the (location) frequency rank. 

\subsubsection{Generation of rank trajectories}
Following earlier assumption, the mobility behavior of an individual is independent from the specific locations the user visits.  For example, those individuals that mostly frequent home and work exhibit  similar behavior, regardless of the actual name of the locations standing for the individual's home and work. In order to analyze and compare the mobility behavior of users, a more abstract and generalized representation of locations is thus needed. 
To deal with this issue, we transform  symbolic trajectories into \emph{rank trajectories}. 
%
%
Specifically, for every summary trajectory $T_s$, we define a  mapping $\rk$ between  the locations  appearing in $T_s$ and the set of frequency ranks  $R=\{1,2,\ldots,r_\text{max}\}$. The ranking is drawn from the number of times relevant locations appear in the trajectory.
If multiple locations have the same frequency, we assign them increasing rank values, so that $\rk$ is a bijective function.
By replacing the symbolic location $l_i$ with the corresponding rank $\rk(l_i)$, we obtain the  rank trajectory: 
$$T_r=(I_1, r_1),\ldots,(I_n, r_n)$$
with $r_i \in R$. Location names are thus replaced by  symbols  representing ordinal  values.  
Rank trajectories share a common set of symbols, in particular the lowest ranks. Therefore, the size of the vocabulary  is significantly  smaller than the original location dictionary.

\begin{example}
Rank trajectory for id=1472043: for brevity, we omit the temporal information and only report the series of ordinal values (delimited by square brackets):
\begin{verbatim}
T=[2 1 2 1 4 2 1 4 2 1 2 1 5 1 2 1 2 8 1 2 1 7
   1 2 1 2 1 9 1 3 1 2 1 2 1 2 1 2 1 3 1 11 2 
   1 2 1 2 1 2 1 2 1 2 1 2 1 2 10 1 2 1 2 1 2 
   1 5 1 2 1 2 12 1 2 3 1 2 1 2 1 2 1 2 1 2 1 
   2 1 2 1 2 1 2 6 1 2 1 2 1 2 1]
\end{verbatim}
The  value $k$ represents the k-th most frequented location in the trajectory. For example, the location of rank 1 is  also referred to in 
literature as \emph{home}, the location of rank 2 as \emph{work}.
\end{example}

\subsubsection{Trajectory splitting}
This operation takes into account the temporal dimension of trajectories. To begin, the observation period is slightly shortened so as to cover an exact number of weeks (a week covers the period from Monday to Sunday).   Rank trajectories are  thus split in smaller trajectories, one per week. Therefore, for any trajectory $T_r$, we obtain a set of \emph{weekly trajectories} $T_w^1,\ldots, T_w^m$, with $m$ the number of weeks.  
The split rule is as follows. For every value $r_i$ in $T_r$ consider the period $I_i$. 
If the starting time of $I_i$  falls into  week $j \in \{1,\ldots,m\}$, $r_i$ is assigned to $T_w^j$.  Weekly trajectories are only defined at symbolic level, while timestamps are ignored, i.e. $$T_w^j=(r_1,\ldots,r_k).$$

\noindent
\begin{example}
\label{ex}
The above trajectory  is split in a series of 10 weekly trajectories $T_w^1,\ldots,T_w^{10}$.
\begin{verbatim}
[2 1 2 1 4 2 1 4 2 1 2 1 5 1]
[2 1 2 8 1 2 1 7]
[1 2 1 2 1 9 1 3 1]
[2 1 2 1 2 1 2 1 3 1 11]
[2 1 2 1 2 1 2 1]
[2 1 2 1 2 1 2 10 1]
[2 1 2 1 2 1 5 1 2 1]
[2 12 1 2 3 1 2 1 2 1 2 1]
[2 1 2 1 2 1 2 1 2 1]
[2 6 1 2 1 2 1 2 1]
\end{verbatim}
A certain regularity can be noticed in the weekly movement, in line with what is reported in literature.  $\diamond$
\end{example}
\subsection{Model training}
 Weekly trajectories are encoded in the form requested  by the Sqn2Vec technique. The Sqn2Vec architecture consists of two feedforward neural networks that are to  be trained separately. One of these networks is trained with the weekly trajectories obtained at the previous step; the second network with the sequential patterns appearing more frequently in the whole dataset given a support threshold and a symbol gap tolerance.  Sequential patterns are defined at dataset scale. Both symbols and sequential patterns are indexed  based on two distinct vocabularies. Hence, for every weekly trajectory, we have two one-hot vectors, one for each training set. The following example illustrates the indexing of sequential patterns.
%
\begin{example}
The sets of  sequential patterns  extracted from the weekly trajectories in Example \ref{ex}  are: 
\begin{verbatim}
SP1={1 5 7 9 65 186 255 259 262 337}
SP2={1 7 8 13 65 66 255 266 312}
SP3={1 6 7 17 65 221 255 260 270 582}
SP4={1 6 7 10 65 221 255 260 263}
SP5={1 7 65 255}
SP6={1 7 24 65 82 255 762}
SP7={1 5 7 65 186 255 259}
SP8={1 6 7 11 64 65 69 221 255 410}
SP9={1 7 65 255}
SP10={1 7 12 65 70 255 439}
\end{verbatim}
Every index corresponds to a  sequential pattern  appearing frequently in the dataset. For example, index 255 corresponds to the two-symbol pattern  (1,2), whilst 259 to (1,5). $\diamond$
\end{example}

Once the input data is organized, the training of the model is performed using the architecture Sqn2Vec-SEP. Sqn2Vec-SEP assigns each trajectory of the two datasets an incremental identifier (the correspondent of the document identifier) and then applies the PV-DBOW technique to train the two models. 
As a result, for each weekly trajectory $i$, we obtain two vector representations in a multi-dimensional embedding space (128 dimensions), $\mathbf{v}_1^{(i)} $ and $\mathbf{v}_2^{(i)} $. The average vector:  $$\mathbf{v}^{(i)}=\frac{\mathbf{v}_1^{(i)} +\mathbf{v}_2^{(i)}}{2} $$  is the final representation of the weekly trajectory $i$, the \emph{weekly embedding}.

\subsection{Generating trajectory embeddings}
\begin{figure}[t]
\centering
\includegraphics[scale=0.22]{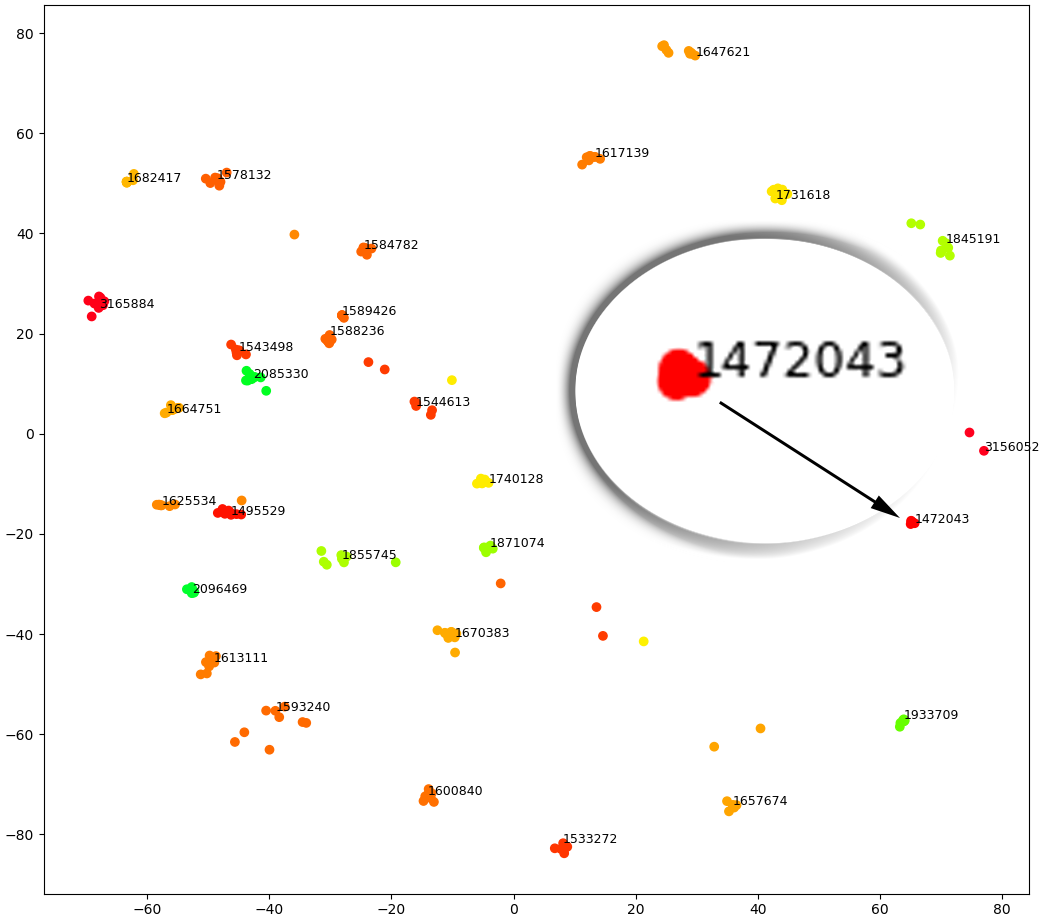}
\caption{Weekly embeddings for a subset of trajectories in 2D space obtained by applying t-SNE for dimensionality reduction. Colors identify the weekly embeddings referring to the same trajectory}
\label{fig:weekly-repr-tsne}
\end{figure}

The next step is to scale up the trajectory representation from  weekly to full  trajectories. Full trajectories cover the whole observation period and are univocally associated to users. To  motivate this operation, consider Figure \ref{fig:weekly-repr-tsne}.

This figure   shows the plot of a small set of weekly embeddings in the two dimensional space obtained by applying the popular t-SNE dimensionality reduction technique. Every point in space represents a weekly trajectory, while the color of the point the full trajectory identifier, i.e. the user. It can be seen that points are clustered by color, namely the weekly trajectories of every user are mapped onto points that are close in space. This is an interesting result, surprisingly in line with the literature on human mobility, according to which the mobility behavior of an individual is similar across the weeks. 
Motivated by this empiric observation, we hypothesize  that the representation of the user's trajectory  can be  computed as centroid of the  set of weekly embeddings. 

Given the trajectory $T_u$ of user $u$  mapped onto $n$ weekly embeddings $\mathbf{v_u^1,v_u^2,... v_u^n} $, we define the \emph{trajectory embedding} of $T_u$ as: $\mathbf{v}_u$, where:
$$\mathbf{v}_u= \frac{1}{n}\sum_i^n \mathbf{v}_u^i.$$

\subsection{Dimensionality reduction}
The embedding space  is reduced to 2 dimensions.  Dimensionality reduction is initially motivated by two factors: the need of supporting the evaluation of \emph{mob2vec} through a visual analysis of the resulting representations; to mitigate the effect of data sparsity due to the curse of dimensionality. Eventually, experiments show that dimensionality reduction is key for an effective representation of full trajectories capable of preserving the behavioral similarity.

For this task, we apply a recent technique, UMAP, that is  proven efficient and scalable \cite{umap2018}.  Built on mathematical foundations, UMAP is a manifold learning technique for dimensionality reduction. 
Similar to other  techniques, e.g. t-SNE, it  preserves the local distance between nearby points against global distances in space.  The authors claim the UMAP algorithm is competitive with t-SNE for visualization quality, and arguably preserves more of the global structure. As a result, we obtain a 2D point for every CDR trajectory, i.e. user.

\section{Experiments: learning process}
We turn to illustrate  the experiments conducted  to evaluate the \emph{mob2vec} pipeline.
In the  next Section  we will focus on the evaluation of the  behavioral similarity. 

\subsection{Overview}


In order to evaluate the goodness of a design choice, whether one solution is preferable to another, we need an evaluation metric. The question is challenging.
To tackle this issue, we propose the following approach: (a) we hypothesize a correlation between the mobility behavior of individuals and the distribution of the location frequency ranks in trajectories; (b) we use such measure of correlation for  evaluating the goodness of embeddings, and thus of the underlying representation model.  

Consider again the example of the two individuals mostly frequenting home and work, while only rarely visiting other locations. These individuals exhibit similar mobility behavior. 
It can also be  seen that the probability distributions of ranks are similar. Therefore for the representation model to be coherent, the trajectory embeddings are expected to be close to each other in space. 
The proposed metric measures the linear correlation coefficient between distances in the embedding space and distances between frequency rank distributions. The metric is detailed next.
It is used to evaluate experimentally the following aspects: 
\begin{enumerate}
\item dimensionality reduction technique;
\item dimension of the embedding space;
\item architecture of the neural network used by Sqn2Vec;
\item rank trajectories vs. summary trajectories;
\item native trajectories vs. summary trajectories.
\end{enumerate}
In more detail, experiments 2 and 3 are for comparison purpose, w.r.t. \cite{Sqn2Vec}. This is also a way to validate the evaluation metric. 
The rest of the experiments are to validate the effectiveness of the key operations on which the learning process is based.

\paragraph{Datasets}
Experiments are conducted on diverse datasets. The original dataset $D_0$ consists of 17000+ CDR trajectories.
In addition, we consider five datasets drawn from $D_0$, and denoted $D_1-D_5$, which are instrumental to the evaluation of different options. The datasets are described  in Table \ref{tab:datasets}, whilst Table \ref{tab:stats} reports summary statistics for each of them, in particular the number of trajectories, the size of the symbol dictionary and the average/max number of symbols per trajectory (trajectory length).
For comparison, we also report in  Table \ref{tabsqn} the statistics  regarding the 8 datasets used to validate Sqn2Vec on real sequential data. It can be seen that the  vocabularies are  comparable in size, while our datasets are significantly larger.

\begin{table}[t]
    \caption{Datasets used in  experiments}
    \label{tab:datasets}
    \begin{adjustbox}{width=\columnwidth,center}
    \begin{tabular}{cp{0.38\textwidth}}
        \toprule
        Dataset & Description\\
        \midrule
        $D_0$ &  Native dataset of 17000+ CDR trajectries\\
        $D_1$ &  Dataset of summary trajectories drawn from $D_0$, by running  SeqScan-d with parameters $N=4, \delta=15'$\\
        $D_2$ & Dataset of weekly trajectories, i.e. summary trajectories ($D_1$) split in weeks\\
        $D_3$ &  Dataset of rank trajectories drawn from summary trajectories ($D_1$)\\
        $D_4$ & Dataset of rank trajectories drawn from weekly trajectories ($D_2$)\\
        $D_5$ & Dataset of rank trajectories drawn from $D_0$ (no summarization) split in weeks\\
        \bottomrule
    \end{tabular}
    \end{adjustbox}
\end{table}
\begin{table}[t]
    \caption{Summary statistics for the datasets used for \emph{mob2vec} evaluation}
    \label{tab:stats}
    \centering
    \begin{adjustbox}{width=\columnwidth,center}
    \begin{tabular}{crrrr}
        \toprule
        Dataset & Trajectories & Symbols & Max. length & Avg. length\\
        \midrule
        $D_0$ & 17241 & 233 &  510 & 107.18\\
        $D_1$ & 151107 & 233 &  78 & 12.22\\
        $D_2$ & 17241 & 95 &  510 & 107.18\\
        $D_3$ & 151107 & 95 &  78 & 12.22\\
        $D_4$ & 152739 & 242 & 12133 & 316.42\\ 
        $D_5$ & 1010238 & 108 &  88 & 9.42\\
        \bottomrule
    \end{tabular}
    \end{adjustbox}
\end{table}
\begin{table}[t]
    \caption{Summary statistics for the datasets used for Sqn2Vec evaluation} 
    \label{tabsqn}
    \centering
    \begin{adjustbox}{width=\columnwidth,center}
    \begin{tabular}{lrrrr}
        \toprule
        Dataset & Sequences & Symbols & Max. length & Avg. length\\
        \midrule
        reuters & 1010 & 6380  & 533 & 93.84\\ 
        aslbu & 424 & 250  & 54 & 13.05\\
        aslgt & 3464 & 94  & 176 & 43.67\\
        auslan2 & 200 & 16 & 18 & 5.53\\
        context & 240 & 94 & 246 & 88.39\\
        pioneer & 160 & 178 &  100 & 40.14\\
        skating & 530 & 82 &  240 & 48.12\\
        unix & 5472 & 1697 &  1400 & 32.34\\
        \midrule
        average & 1437 & 1099 & 346 & 45.64\\
        \bottomrule
    \end{tabular}
    \end{adjustbox}
\end{table}

\paragraph{Hardware and software settings.} For the experiments we have used the following settings.
%
Hardware: HP ProLiant ML350 G6, with Windows 10 LTSC 1809, equipped with: CPU Intel Xeon X5675 6core/12threads, 120GB DDRIII ECC RAM, 500GB SSD. 
%
Relevant software libraries and platforms: 
PostGIS; 
PyCharm 2019; 
scikit-learn 0.22.1; 
umap-learn 0.3.10;
gensim 3.8.0;
github.com/nphdang/\-Sqn2Vec post commit 056cc53 (Sqn2Vec).

\subsection{Evaluation metric}

Consider two rank trajectories $T_1, T_2$.
We  define: 
\begin{itemize}
\item The distance between the corresponding trajectory embeddings $e_1, e_2$ is the Euclidean distance $d(e_1, e_2)$ on the 2D plane. 
\item The distance between the probability distributions $g_1, g_2$ of the rank symbols is the Jensen-Shannon distance $D_{JS}(g_1,g_2)$, i.e. the square root of the Jensen-Shannon divergence \cite{jensenShannon}.
$D_{JS}$ is based on the Kullback–Leibler divergence ($D_{KL}$) and have some useful properties: in particular, is a metric and takes values ranging between 0 and 1.
$D_{JS}$ is defined as follows: 
\begin{gather*}
D_{JS}(g_1,g_2) = \sqrt{\frac{1}{2}D_{KL}(g_1|M) + \frac{1}{2}D_{KL}(g_2|M)}\\
\text{where }M = \frac{1}{2}(g_1+g_2).
\end{gather*}
\end{itemize}
Given a dataset of rank trajectories, we compute for every pair of trajectories $i, j$, the point of coordinates $d(e_i, e_j)$ and $D_{JS}(g_i,g_j)$, respectively, hence  the linear correlation between the two distances using the Pearson linear correlation coefficient $r \in [-1,+1]$.
The evidence of a strong positive correlation (i.e.  $0.5 \leq r \leq 1 $) is interpreted as goodness of the trajectory embeddings (based on the given hypothesis). 

\subsection{ Experiment 1:   dimensionality reduction}
The goal is to evaluate  to what extent the  dimensional reduction of the trajectory embeddings is beneficial for the learning process. 
%
The method is  as follows. We start evaluating the goodness of embeddings when the space dimensions are not reduced,  next we analyze various options for the reduction of dimensionality. The source dataset  is $D_4$, i.e. rank trajectories split in weeks. The dimension of the trajectory embeddings is set to  128. The experiments are: 
%
\paragraph{Exp. 1.1}
For generality, we consider two distance functions in the embedding space, the cosine similarity and the Euclidean distance.
Table \ref{tab:exp1} reports  the coefficient $r$ resulting in the two cases. It can be seen that there is no evidence of strong linear  correlation. 
\begin{table}[h]
    \caption{No dimensionality reduction} 
    \label{tab:exp1}
    \centering
    \begin{tabular}{ccc} 
        \toprule
        Space dimensions & Distance & $r$\\
        \midrule
        128 & cosine & 0.0508\\ 
        128 & Euclidean & 0.3628\\
        \bottomrule
    \end{tabular}
\end{table}

\paragraph{Exp. 1.2}
Seven dimensionality reduction techniques have been selected for evaluation and comparison. Specifically: 
\begin{itemize}
\item Principal Component Analysis (PCA);
\item Isometric Mapping (IsoMap);
\item Multi-dimensional Scaling (MDS);
\item t-distributed Stochastic Neighbor Embedding (t-SNE);
\item Locally Linear Embedding (LLE);
\item Spectral Embedding;
\item Uniform Manifold Approximation and Projection for Dimension Reduction (UMAP).
\end{itemize}

\begin{figure}[t]
\centering
\subfigure[]{\includegraphics[width=3.3cm]{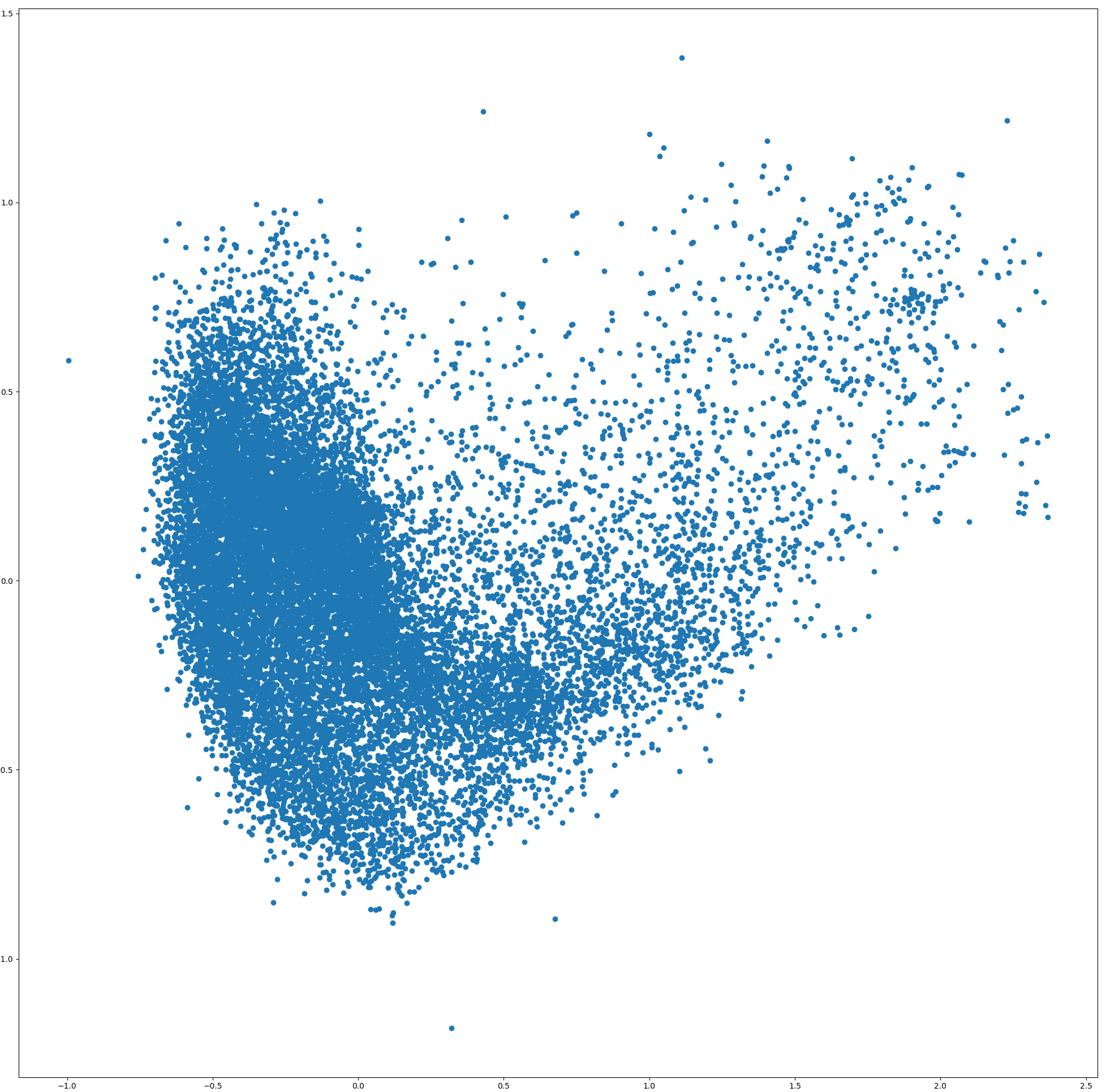}}
\subfigure[]{\includegraphics[width=3.3cm]{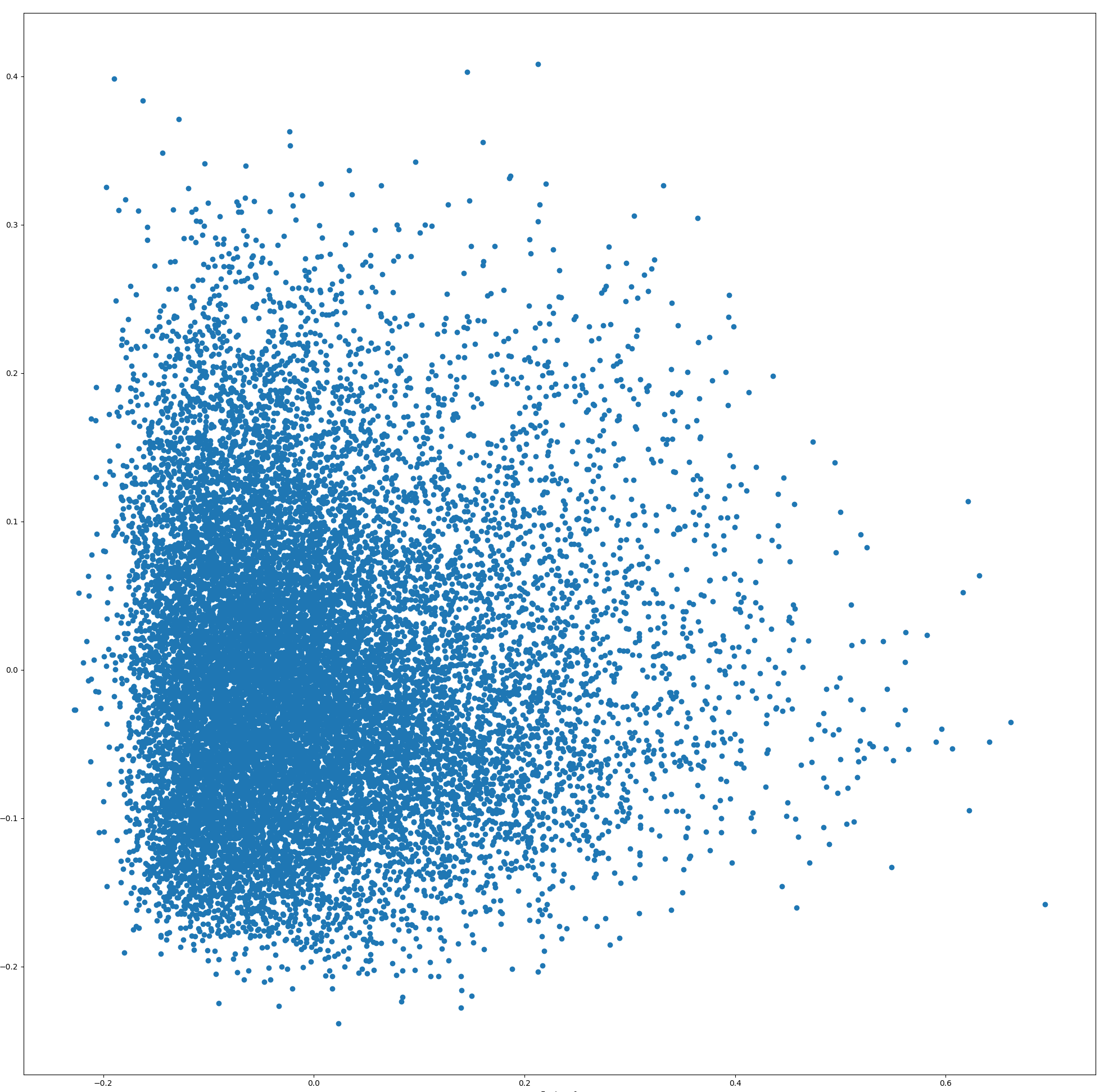}}
\subfigure[]{\includegraphics[width=3.3cm]{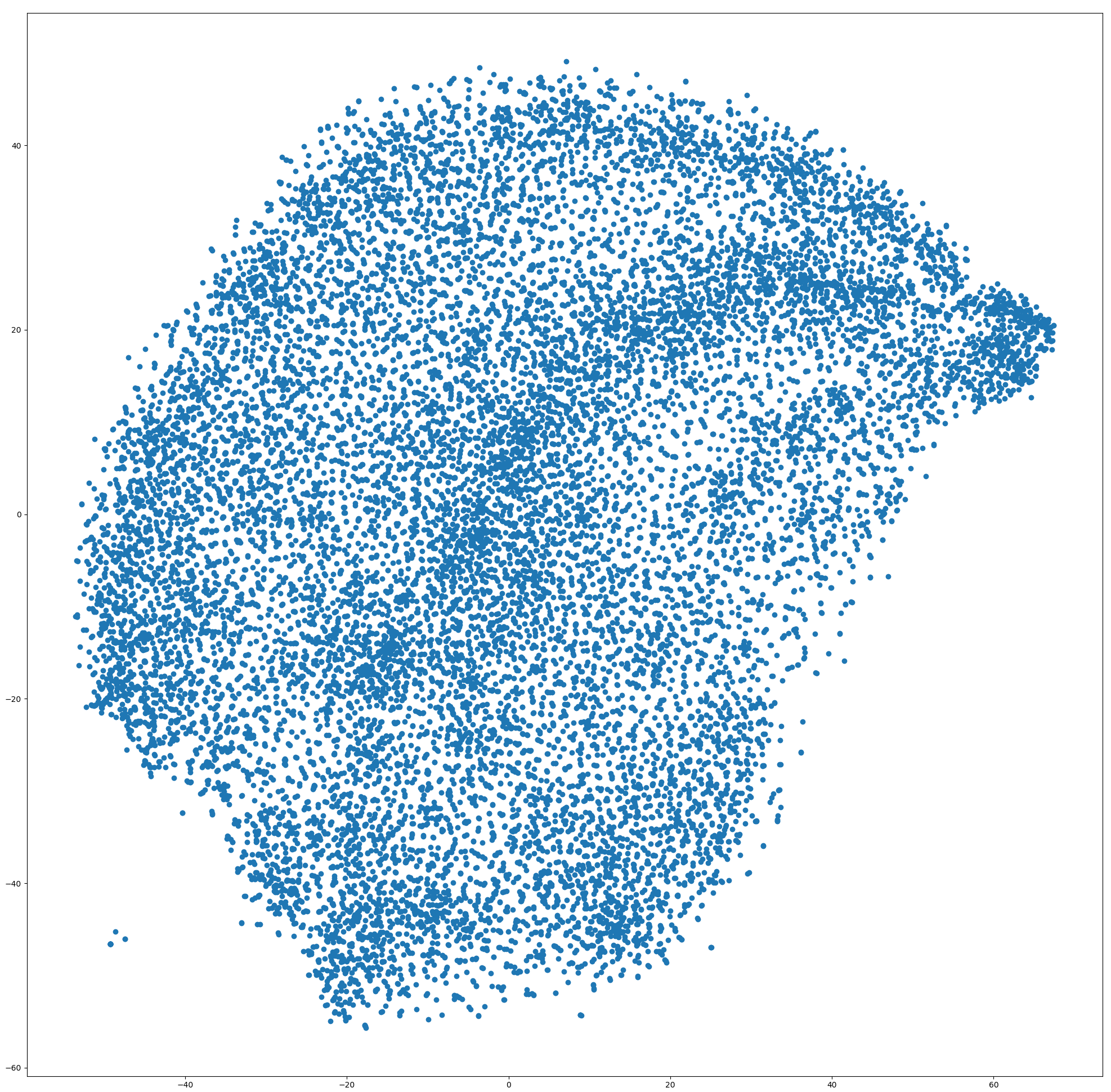}}
\subfigure[]{\includegraphics[width=3.3cm]{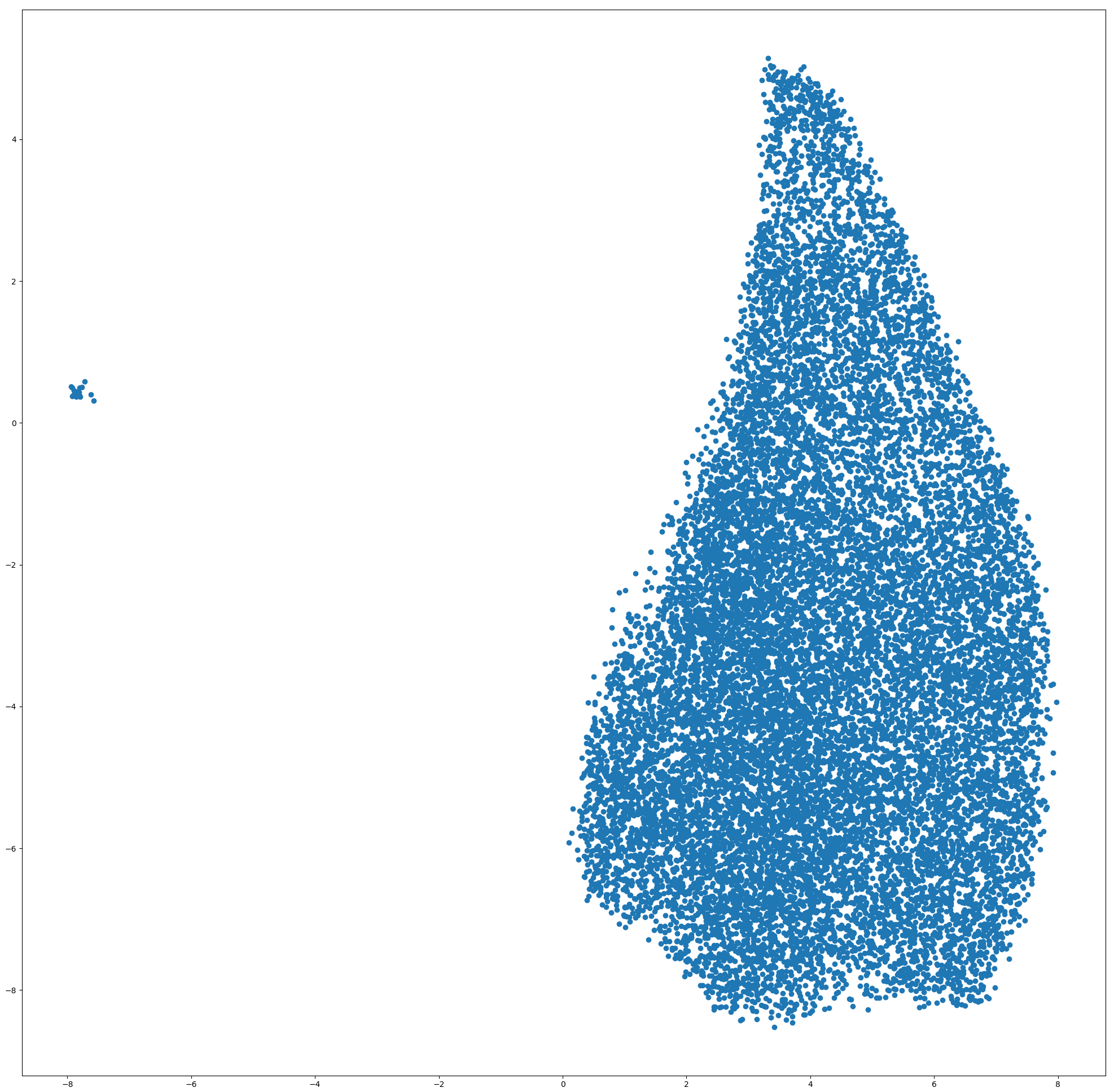}}
\caption{Examples of dimensional reduction of trajectory embeddings: (a) Isometric Mapping; (b) Principal Component Analysis; (c) t-SNE; (d) UMAP. In all nonlinear methods (a, c, d) the axes have no specific meaning: any rotation produces an equally valid embedding.}
\label{isomap-mds}
\end{figure}

\begin{figure}[t]
\centering
\includegraphics[width=6.5cm]{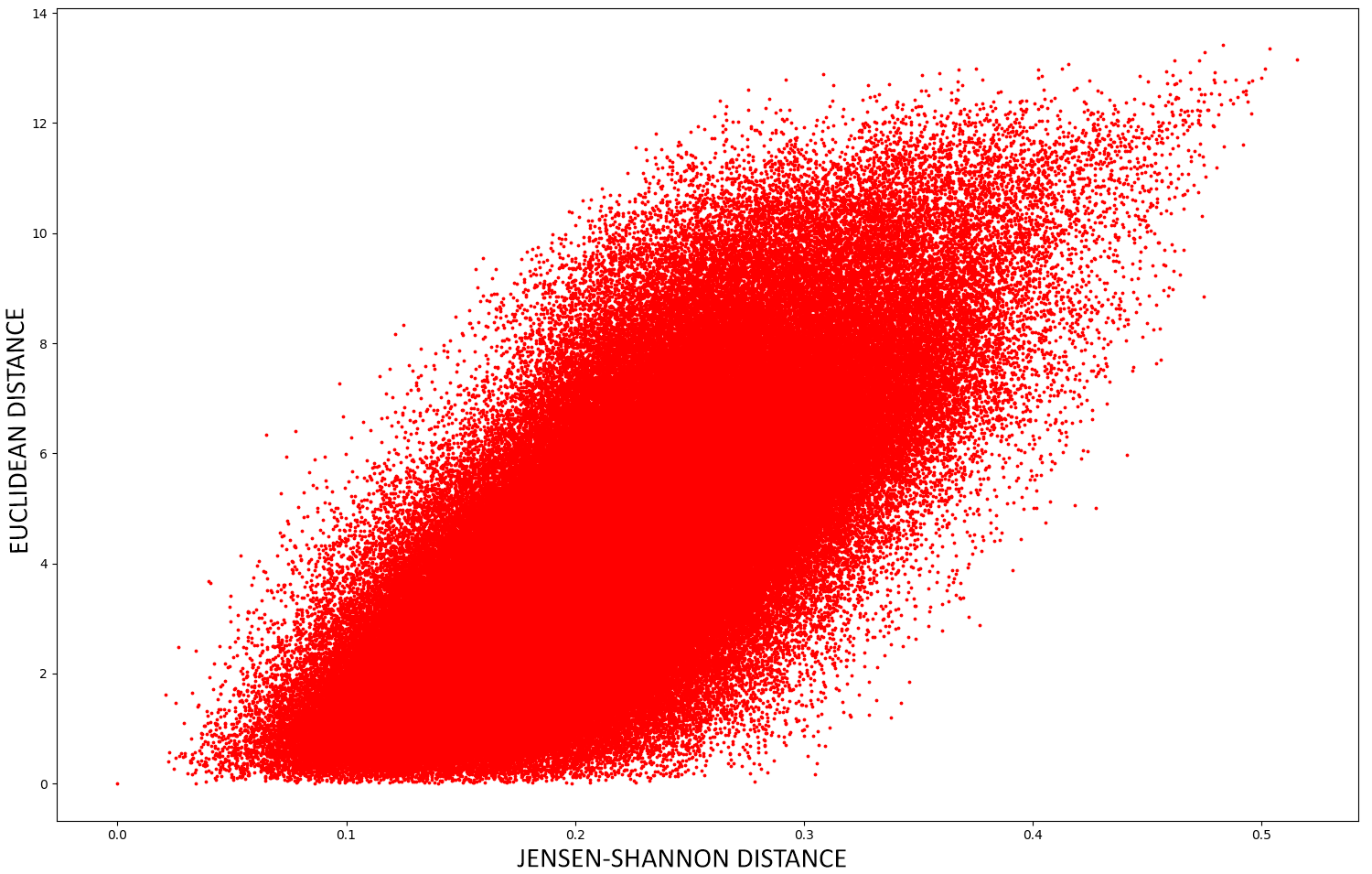}
\caption{Scatter plot of points representing pairs of rank trajectories: the y-axis reports the Euclidean distance of the embeddings computed with UMAP, the x-axis the Jensen-Shannon distance between symbol distributions} 
\label{figpcorr800}
\end{figure}

The evaluation considers both the coefficient $r$ and the computational efficiency of the technique (run time).  The results are
computed by extracting the vectors at the end of the training.  
They are shown in Table \ref{tab:exp12}, while a few exemplifying plots are reported in Figure \ref{isomap-mds}. It can be seen that  PCA and UMAP are the most efficient and also the value of $r$ is high, i.e. the trajectory embeddings are coherent with symbol distribution. 
Unlike PCA, and in the same line of t-SNE applied for visualization purposes in \cite{Sqn2Vec},  UMAP is a non-linear transformation technique. 
UMAP is chosen as part of the methodology and used in the rest of the experiments.  Figure \ref{figpcorr800} illustrates the meaning of the linear correlation coefficient computed for UMAP.

\begin{table}[h!]
  \caption{Evaluation of dimensional reduction techniques} 
    \label{tab:exp12}
    \centering
    \begin{tabular}{lrr}
        \toprule
        Technique & $r$ & Runtime [s]\\
        \midrule
        PCA & 0.7100 & 1\\ 
        IsoMap & 0.6376 & 433\\
        MDS & 0.5075 & 1696\\
        t-SNE & 0.6487 & 277\\
        LLE & 0.6285 & 529\\
        Spectral & 0.5493 & 473\\
        UMAP & 0.7302 & 69 \\
     \bottomrule
    \end{tabular}
\end{table}

\subsection{Experiment 2: space dimensions} We evaluate the impact of the embedding space dimension over the correlation $r$. A strong correlation $r$ can be observed in higher dimensional spaces. Table \ref{tab:exp2} 
reports the value of $r$ for spaces of 64,128 and 256 dimensions, respectively. As in Sqn2Vec, we choose an embedding space of 128 dimensions. 

\begin{table}[h!]
    \caption{Impact of the number of dimensions} 
    \label{tab:exp2}
    \centering
    \begin{tabular}{cc} 
        \toprule
        Space dimensions & $r$\\ 
        \midrule
        \ 64 & 0.7224\\ 
        128 & 0.7302\\
        256 & 0.7176\\
        \bottomrule
    \end{tabular}
\end{table}

\subsection{Experiment 3: PV-DM vs. PV-DBOW}

The Sqn2Vec technique relies on the architecture PV-DBOW, which in \cite{Sqn2Vec} is proven more effective than PV-DM.  We recall that  PV-DM is the context-aware neural network architecture defined in \cite{ParagraphVector1}. In this experiment, we evaluate whether we can reach the same conclusion using our evaluation metric for comparison. We also compare the efficiency of the training process. The parameter indicating the width of the context window (PV-DM)  is set to 5 symbols.  Results are reported in Table \ref{tab:exp3}. It is shown that  PV-DBOW has a better performance, in line with  the literature. The  evaluation metric is  coherent.  
\begin{table}[h!]
    \caption{Experiment 3: PV-DBOW vs. PV-DM} 
    \label{tab:exp3}
    \centering
    \begin{tabular}{lcr}
        \toprule
        NN Architecture & $r$ & Training duration [s]\\ 
        \midrule
        PV-DM & 0.5704 & 22412\\ 
        PV-DBOW & 0.7302 & 9802\\
        \bottomrule
    \end{tabular}
\end{table}

\subsection{Experiment 4: location vs. rank} A  key step  of the learning process is the generalization of  summary trajectories into learning  rank trajectories.  In this experiment,  we consider two different source datasets, i.e. $D_2$ and $D_4$, containing  summary and rank trajectories  respectively. Two sets of trajectory embeddings are then generated  and finally evaluated using the $r$ coefficient. Table \ref{tab:exp4} reports the results. It is evident that rank trajectories are key for the learning of behavioral models. 
\begin{table}[h!]
    \caption{Experiment 4: location vs. rank symbols} 
    \label{tab:exp4}
    \centering
    \begin{tabular}{lc}
        \toprule
        Trajectory type & $r$\\
        \midrule
        Summary trajectories ($D_2$) & 0.0036\\ 
        Rank trajectories ($D_4$) & 0.7302\\
        \bottomrule
    \end{tabular}
\end{table}

\subsection{Experiment 5: native vs. summary data} 
Summary trajectories contain \emph{relevant} locations extracted from native trajectories using the SeqScan-d technique. The main purpose of this operation, called summarization, is to remove noise and insignificant locations. In this experiment, we evaluate the impact of summarization.  We consider the  dataset $D_5$ of  native trajectories and the data set $D_4$ of summary trajectories, and  evaluate the coefficient $r$ in two  embedding spaces, of 128 and 256 dimensions, respectively. Table \ref{tab:exp5} reports the evaluation. It can be seen that the coefficient $r$ and thus the quality of embeddings is definitely higher when summary trajectories are used.

\begin{table}[h!]
    \caption{Experiment 5: native vs. summary trajectories} 
    \label{tab:exp5}
    \centering
    \begin{adjustbox}{width=\columnwidth,center}
    \begin{tabular}{lcc}
        \toprule
        Trajectory type & Space dimensions & $r$\\ 
        \midrule
        Native trajectories ($D_5$) & 128 & 0.0615\\ 
        Summary trajectories ($D_4$) & 128 & 0.7302\\ 
        Native trajectories ($D_5$) & 256 & 0.0348\\ 
        Summary trajectories ($D_4$) & 256 & 0.7176\\
        \bottomrule
    \end{tabular}
    \end{adjustbox}
\end{table}


\section{Experiments: behavioral similarity}
\begin{figure}[t]
\centering
\includegraphics[scale=0.16]{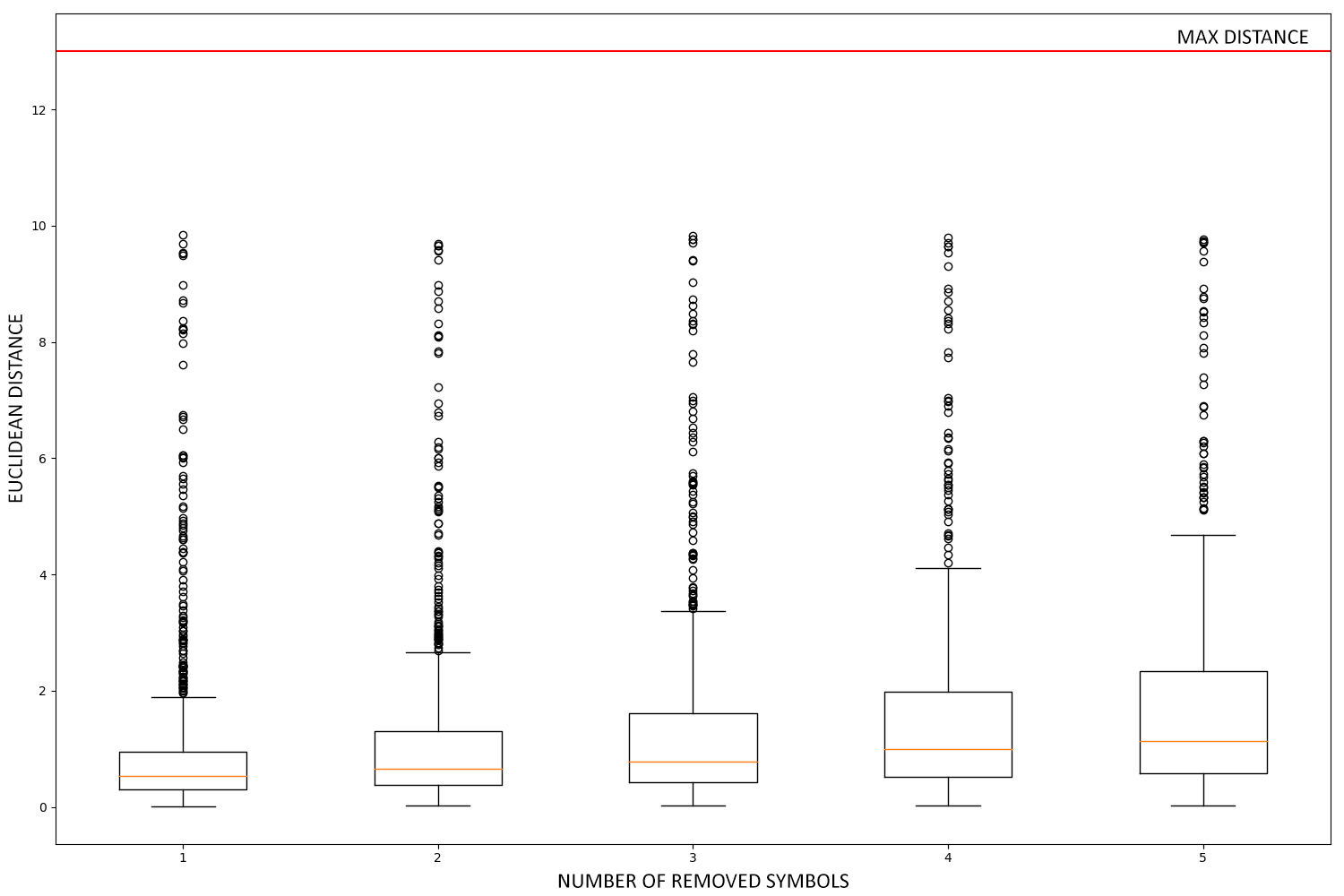}
\caption{Boxplot reporting the distribution  of the Euclidean distance between the embeddings of  source trajectories  and similar trajectories for increasing values of removed ranks.}
\label{boxplot}
\end{figure}

 We turn to evaluate the capability of preserving behavioral similarity. The goal is to create trajectories similar to those  defined in the  training set, and then evaluate the distance among the respective embeddings. 
 Following the methodology, we should  find that those embeddings are in proximity.
 The first non-trivial question is how to specify trajectories, and more specifically rank trajectories, that are \emph{similar} to those of the training set.  To tackle the problem, an approach  \cite{t2vec} is to apply down-sampling and distortion.  
 In that project, however, trajectories are spatial trajectories acquired at high sampling rate, while rank trajectories are sequences of ordinal values corresponding to sparse locations. We thus propose a different approach.
The idea is to remove from the rank trajectories the values occurring in the sequence that are less significant because they correspond to locations that are rarely frequented (i.e. the ranks of higher value). It can be seen that the removal of the rank value $n$ from a trajectory does not alter the lowest ranks $1,2,\ldots,n-1$, that is, the relevant characteristics of the mobility behavior of an individual are not altered. In this sense, this form of down-sampling generates rank trajectories that are slightly different from the original trajectories and thus are \emph{similar}. 
%
The experiment is as follows.
We randomly select 1000 trajectories from the source dataset $D_4$ (source trajectories). For every trajectory, we create 5 \emph{similar} trajectories by removing the $k$ less significant rank values for $k \in [1,5]$. The degree of similarity  thus decreases for increasing values of $k$. Based on these  data,  we use the functionality of representation inference to generate the embeddings for the similar trajectories. The boxplot in Figure \ref{boxplot} describes the variation in distance between similar trajectories and  source trajectories for increasing value of $k$. The red line in the figure reports the maximum distance between two embedded points. It can be seen that: (a) the distance measures are relatively small; (b) on average, the distance increases monotonically with the value of $k$.  
These results are coherent with the methodological framework. 

\begin{figure}[t]
\centering
\includegraphics[width=7.9cm]{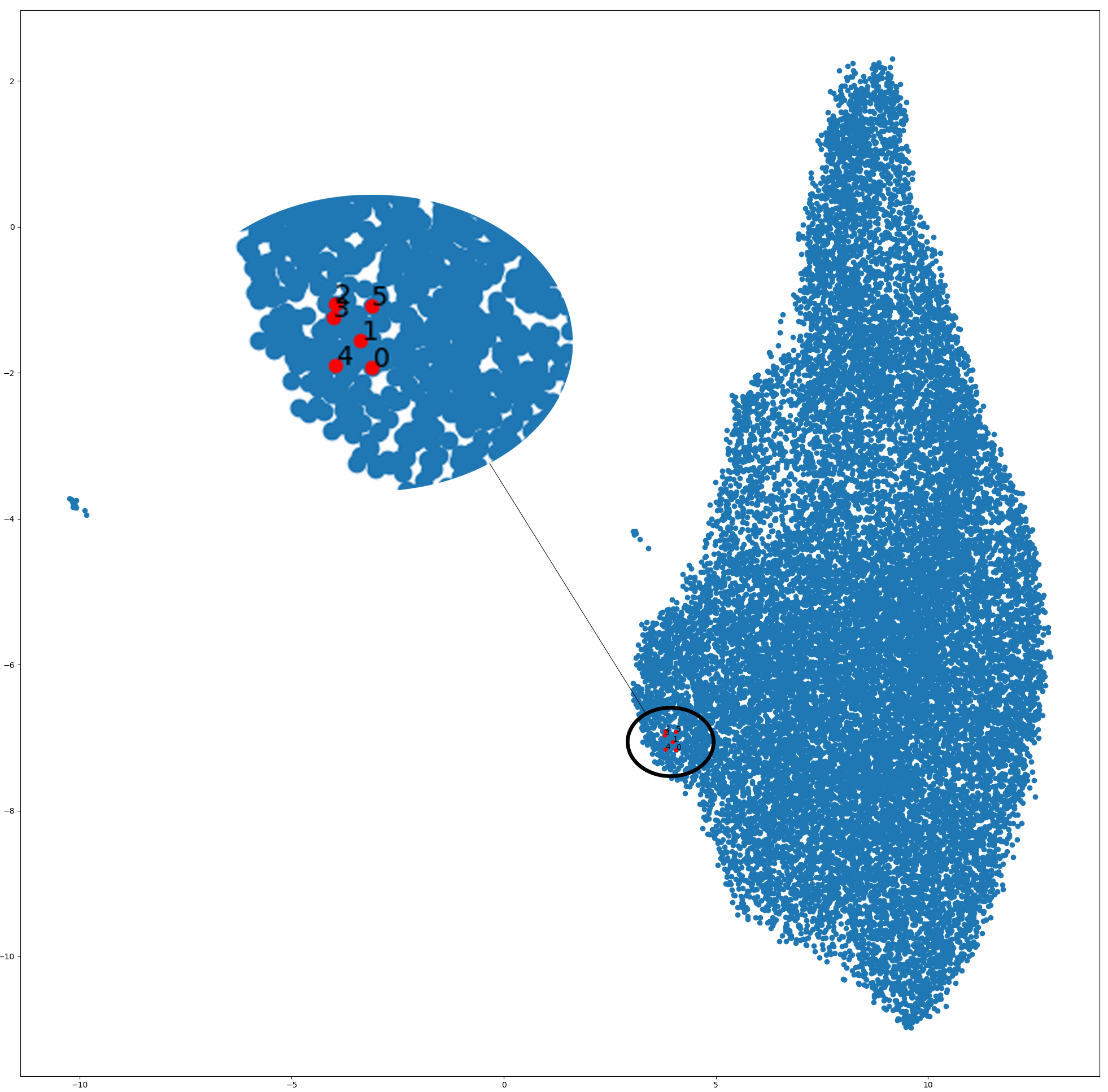}
\caption{Visual analysis of embeddings: source trajectory (point 0) and the five similar trajectories (points 1 to 5).}
\label{5trajs-2d}
\end{figure}

Visual evidence of the previous result is given in  Figure \ref{5trajs-2d}. In particular, the figure  highlights the representation of a specific trajectory in the 2D embedding space (point with index 0), close to the embedded points of the five similar trajectories (indices 1 to 5). 

\begin{figure}[t]
\centering
\includegraphics[width=7.9cm]{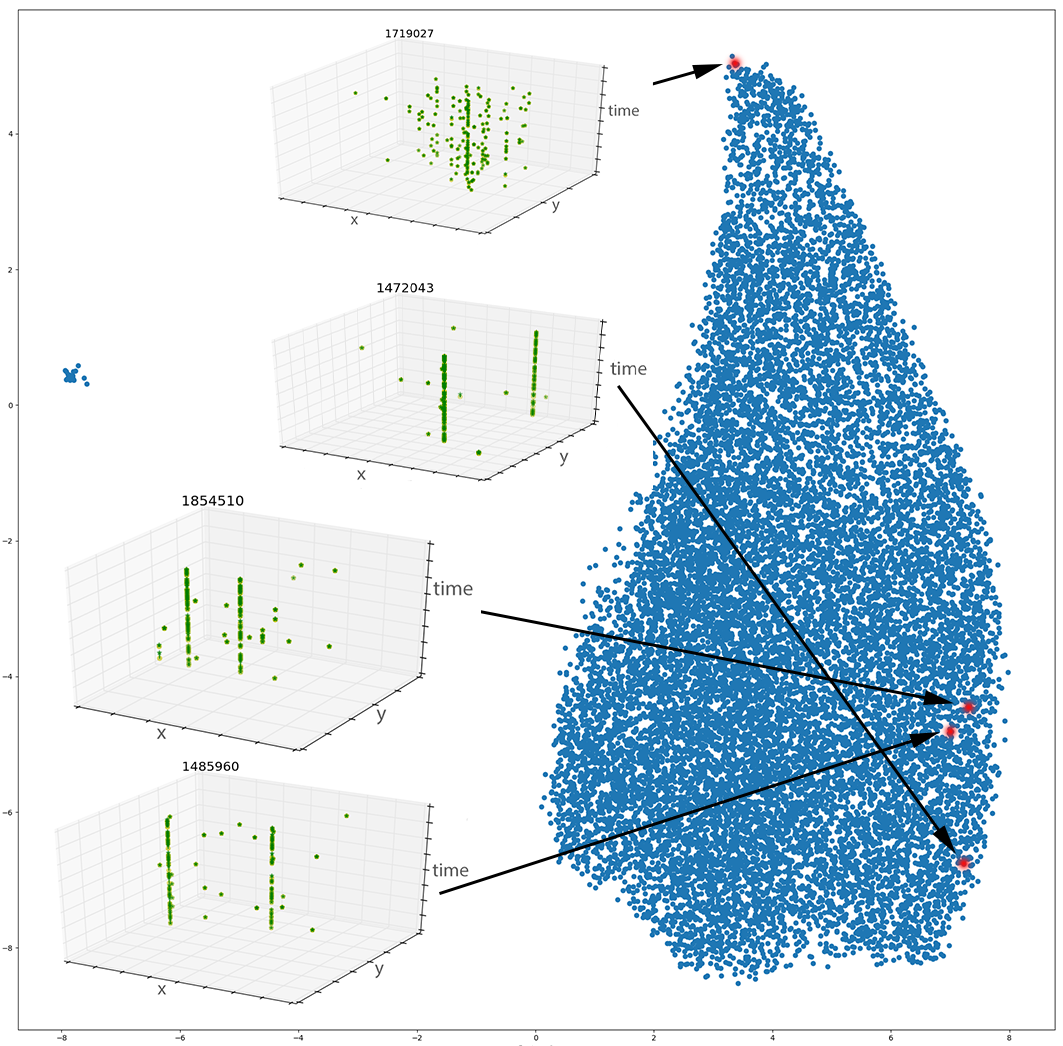}
\caption{The embeddings  corresponding to the four trajectories represented in the spatio-temporal space.}
\label{fig:last}
\end{figure}

Further insights are provided by Figure \ref{fig:last}. The figure highlights the embedded points representing the four summary trajectories displayed in the spatio-temporal coordinated space. In this case the trajectories belong to the source dataset $D_4$. It can be seen that the distance  reflects the diversity of the trajectories. In particular the furthest point represents a trajectory that appears substantially different from the other trajectories.

\section{Conclusions}
In this paper, we have presented the \emph{mob2vec} 
framework for mapping CDR trajectories into vector representations capable of
preserving mobility behavior similarity. The approach builds on 
Sqn2Vec for the learning of sequence representations. Yet, Sqn2Vec does not provide any support for time, neither for noisy data and behavioral similarity, which instead is the major contribution of this work. We have been able to show the effectiveness of key choices: the generalization of movement in rank trajectories; noise removal;  
the aggregation of temporally-bounded vectors; the evaluation of  the embedding quality; 
the UMAP dimensionality reduction of the embedding space. 
The result is exciting. We plan to apply the technique to other datasets, in order  to 
consolidate the results. Yet, the major challenge is to provide a theoretical explanation, corroborating the empirical achievements.

\paragraph{Acknowledgments}
The research carried out by M. L. Damiani, F. Hachem  and M. Rossini is partially supported by the Italian government via the NG-UWB project (MIUR PRIN 2017).

\bibliographystyle{abbrv}
  \footnotesize\bibliography{main.bib}

\begin{thebibliography}{10}

\bibitem{barbosa2018}
H.~Barbosa, M.~Barthelemy, G.~Ghoshal, C.~R. James, M.~Lenormand, T.~Louail,
  R.~Menezes, J.~J. Ramasco, F.~Simini, and M.~Tomasini.
\newblock Human mobility: Models and applications.
\newblock {\em Physics Reports}, 734:1 -- 74, 2018.

\bibitem{Bengio2013}
Y.~{Bengio}, A.~{Courville}, and P.~{Vincent}.
\newblock Representation learning: A review and new perspectives.
\newblock {\em IEEE Transactions on Pattern Analysis and Machine Intelligence},
  35(8):1798--1828, 2013.

\bibitem{lcss2018}
K.~Bringman and M.~K{\"u}nnemann.
\newblock Multivariate fine-grained complexity of longest common subsequence.
\newblock In {\em Proc. of the 29th ACM-SIAM Symposium on Discrete Algorithms},
  pages 1216--1235, 2018.

\bibitem{sigspatial2019}
D.~K. Chandra, P.~Wang, J.~Leopold, and Y.~Fu.
\newblock Collective representation learning on spatiotemporal heterogeneous
  information networks.
\newblock In {\em Proc. of the 27th ACM SIGSPATIAL}, 2019.

\bibitem{2013physica}
B.~Csaji, A.~Browet, V.~Traag, J.~Delvenne, E.~Huens, P.~Van~Dooren,
  Z.~Smoreda, and V.~D. Blondel.
\newblock Exploring the mobility of mobile phone users.
\newblock {\em Physica A: Statistical Mechanics and its Applications},
  392(6):1459 -- 1473, 2013.

\bibitem{ParagraphVector2}
A.~M. Dai, C.~Olah, and Q.~Le.
\newblock Document embedding with paragraph vectors.
\newblock {\em arXiv preprint arXiv:1507.07998}, 2015.

\bibitem{Damiani2017}
M.~L. Damiani and F.~Hachem.
\newblock Segmentation techniques for the summarization of individual mobility
  data.
\newblock {\em Wiley Interdisciplinary Reviews: Data Mining and Knowledge
  Discovery}, 7(6), 2017.

\bibitem{DamianiHIRMC18}
M.~L. Damiani, F.~Hachem, H.~Issa, N.~Ranc, P.~Moorcroft, and F.~Cagnacci.
\newblock Cluster-based trajectory segmentation with local noise.
\newblock {\em Data Mining and Knowledge Discovery}, 32(4):1017--1055, 2018.

\bibitem{DamianiHQG19}
M.~L. Damiani, F.~Hachem, C.~Quadri, and S.~Gaito.
\newblock Location relevance and diversity in symbolic trajectories with
  application to telco data.
\newblock In {\em Proc. of the 16th International Symposium on Spatial and
  Temporal Databases, {SSTD}}, pages 41--50, 2019.

\bibitem{tsas2020}
M.~L. Damiani, F.~Hachem, C.~Quadri, M.~Rossini, and S.~Gaito.
\newblock On location relevance and diversity in human mobility data.
\newblock {\em ACM TSAS}, In press.

\bibitem{fournier2017frequentmining}
P.~Fournier-Viger, J.~C.-W. Lin, R.~U. Kiran, Y.-S. Koh, and R.~Thomas.
\newblock A survey of sequential pattern mining.
\newblock {\em Data Science and Pattern Recognition}, 1(1):54--77, 2017.

\bibitem{jensenShannon}
B.~Fuglede and F.~Topsoe.
\newblock {Jensen-Shannon} divergence and {Hilbert} space embedding.
\newblock In {\em Proc. of the IEEE International Symposium on Information
  Theory}, 2004.

\bibitem{barabasi2008}
M.~C. Gonzales, C.~Hidalgo, and A.~Barabasi.
\newblock Understanding individual individual human mobility patterns.
\newblock {\em Nature}, 453(7196):779--782, 2008.

\bibitem{tsas2015}
R.~H. G{\"u}ting, F.~Vald{\'e}s, and M.~L. Damiani.
\newblock {Symbolic trajectories}.
\newblock {\em Trans. Spatial Algorithms and Systems}, 1(2):7:1--7:51, 2015.

\bibitem{ParagraphVector1}
Q.~Le and T.~Mikolov.
\newblock Distributed representations of sentences and documents.
\newblock In {\em Proc. of the 31st International Conference on Machine
  Learning (ICML)}, volume~32, page II–1188–II–1196. JMLR.org, 2014.

\bibitem{t2vec}
X.~{Li}, K.~{Zhao}, G.~{Cong}, C.~S. {Jensen}, and W.~{Wei}.
\newblock Deep representation learning for trajectory similarity computation.
\newblock In {\em Proc. of IEEE 34th International Conference on Data
  Engineering (ICDE)}, 2018.

\bibitem{umap2018}
L.~McInnes, J.~Healy, and J.~Melville.
\newblock Umap: Uniform manifold approximation and projection for dimension
  reduction.
\newblock {\em arXiv preprint arXiv:1802.03426}, 2018.

\bibitem{word2vec2}
T.~Mikolov, I.~Sutskever, K.~Chen, and J.~Dean.
\newblock Distributed representations of words and phrases and their
  compositionality.
\newblock In {\em Advances in Neural Information Processing Systems 26}. 2013.

\bibitem{mooney2013frequentmining}
C.~Mooney and J.~Roddick.
\newblock Sequential pattern mining--approaches and algorithms.
\newblock {\em ACM Computing Surveys (CSUR)}, 45(2):1--39, 2013.

\bibitem{Sqn2Vec}
D.~Nguyen, W.~Luo, T.-D. Nguyen, S.~Venkatesh, and D.~Phung.
\newblock {Sqn2Vec}: Learning sequence representation via sequential patterns
  with a gap constraint.
\newblock In {\em Machine Learning and Knowledge Discovery in Databases}, pages
  569--584. Springer International Publishing, 2019.

\bibitem{2016sab}
M.~Papandrea, K.~Keramat, M.~Zignani, S.~Gaito, S.~Giordano, and G.~Rossi.
\newblock On the properties of human mobility.
\newblock {\em Computer Communications}, 87:19--36, 2016.

\bibitem{pappalardo2015gyration}
L.~Pappalardo, F.~Simini, S.~Rinzivillo, D.~Pedreschi, F.~Giannotti, and
  A.~Barab{\'a}si.
\newblock Returners and explorers dichotomy in human mobility.
\newblock {\em Nature communications}, 6(1):1--8, 2015.

\bibitem{2018sab}
C.~{Quadri}, M.~{Zignani}, S.~{Gaito}, and G.~P. {Rossi}.
\newblock On non-routine places in urban human mobility.
\newblock In {\em Proc. of IEEE 5th International Conference on Data Science
  and Advanced Analytics (DSAA)}, 2018.

\bibitem{doc2vecImplementation}
R.~{\v R}eh{\r u}{\v r}ek and P.~Sojka.
\newblock Software framework for topic modelling with large corpora.
\newblock In {\em Proc. of the LREC Workshop on New Challenges for NLP
  Frameworks}, pages 45--50. ELRA, 2010.

\bibitem{barabasi2010}
C.~Song, T.~Koren, P.~Wang, and A.~Barabási.
\newblock Modelling the scaling properties of human mobility.
\newblock {\em Nature Physics}, 6:818–823, 2010.

\bibitem{t2vec-explained}
S.~Taghizadeh, A.~Elekes, M.~Schäler, and K.~Böhm.
\newblock How meaningful are similarities in deep trajectory representations?
\newblock {\em Information Systems - in press}, 2019.

\bibitem{kdd2019}
P.~Wang, Y.~Fu, H.~Xiong, and X.~Li.
\newblock Adversarial substructured representation learning for mobile user
  profiling.
\newblock In {\em Proc. of ACM SIGKDD (KDD'19)}, 2019.

\bibitem{kdd2018}
P.~Wang, Y.~Fu, J.~Zhang, P.~Wang, Y.~Zheng, and C.~Aggarwal.
\newblock You are how you drive: Peer and temporal-aware representation
  learning for driving behavior analysis.
\newblock In {\em Proc. of ACM SIGKDD (KDD'18)}, 2018.

\bibitem{sigspatial2017}
B.~Yan, K.~Janowicz, G.~Mai, and S.~Gao.
\newblock From itdl to place2vec: Reasoning about place type similarity and
  relatedness by learning embeddings from augmented spatial contexts.
\newblock In {\em Proc. of the 25th ACM SIGSPATIAL}, 2017.

\end{thebibliography}

\end{document}